\documentclass[letterpaper, 10 pt, conference]{ieeeconf}  % Comment this line out if you need a4paper

\IEEEoverridecommandlockouts                              % This command is only needed if 
\usepackage{amsmath,amssymb,amsfonts}
\usepackage{algorithmic}
\usepackage{graphicx}
\usepackage{textcomp}
\usepackage{xcolor}
\usepackage{bm}
\usepackage{amsmath} 
\usepackage{subfigure}
\usepackage{tabularx}
\usepackage{multirow}
\usepackage{diagbox}
\usepackage{threeparttable}
\usepackage{txfonts}
\usepackage{subfigure}
\usepackage{algorithm}
\usepackage{algorithmic} 
\usepackage{mathrsfs}
\usepackage{booktabs}
\usepackage{marvosym}
\usepackage{stfloats}
\let\labelindent\relax
\usepackage{enumitem}
\usepackage{siunitx} 
\usepackage{caption}
\usepackage{url}
\usepackage{xcolor}
\usepackage{authblk}
\usepackage{cite}

\title{\LARGE \bf
Dexterous Grasping with Real-World Robotic Reinforcement Learning
}

\author{
Dongchi Huang, Tianle Zhang, Yihang Li, Ling Zhao, Jiayi Li, Zhirui Fang, \\ Chunhe Xia, and Xiaodong He
\thanks{Dongchi Huang and Chunhe Xia are affiliated with Beihang University, Beijing, China. Email: {\tt\small \{zy2306316,xch\}@buaa.edu.cn}.}
\thanks{Tianle Zhang, Yihang Li, Ling Zhao, Lusong Li, and Xiaodong He are associated with JD Explore Academy, Beijing, China. Email: {\tt\small tianle-zhang@outlook.com; \{liyihang18, zhaolin120, hexiaodong\}@jd.com}.}
\thanks{Jiayi Li is affiliated with Beijing Jiaotong University, Beijing, China. Email: {\tt\small lijiayi140@jd.com}.}
\thanks{Zhirui Fang is associated with Tsinghua University, Beijing, China. Email: {\tt\small fangzhirui.2@jd.com}.}
}
\begin{document}

\maketitle
\thispagestyle{empty}
\pagestyle{empty}

%%%%%%%%%%%%%%%%%%%%%%%%%%%%%%%%%%%%%%%%%%%%%%%%%%%%%%%%%%%%%%%%%%%%%%%%%%%%%%%%
\begin{abstract}
Dexterous grasping in the real world presents a fundamental and significant challenge for robot learning. The ability to employ affordance-aware poses to grasp objects with diverse geometries and properties in arbitrary scenarios is essential for general-purpose robots. However, existing research predominantly addresses dexterous grasping problems within simulators, which encounter difficulties when applied in real-world environments due to the domain gap between reality and simulation. This limitation hinders their generalizability and practicality in real-world applications. 
In this paper, we present DexGraspRL, a reinforcement learning (RL) framework that directly trains robots in real-world environments to acquire dexterous grasping skills. Specifically, DexGraspRL consists of two stages: \textit{(i)} a pretraining stage that pretrains the policy using imitation learning (IL) with a limited set of expert demonstrations; \textit{(ii)} a fine-tuning stage that refines the policy through direct RL in real-world scenarios.
To mitigate the catastrophic forgetting phenomenon arising from the distribution shift between demonstrations and real-world environments, we design a regularization term that balances the exploitation of RL with the preservation of the pretrained policy.
Our experiments with real-world tasks demonstrate that DexGraspRL successfully accomplishes diverse dexterous grasping tasks, achieving an average success rate of nearly 92\%. Furthermore, by fine-tuning with RL, our method uncovers novel policies, surpassing the IL policy with a 23\% reduction in average cycle time.
% Our code and empirical results are available in \url{https://hggforget.github.io/iborl.github.io/}.
\end{abstract}

%%%%%%%%%%%%%%%%%%%%%%%%%%%%%%%%%%%%%%%%%%%%%%%%%%%%%%%%%%%%%%%%%%%%%%%%%%%%%%%%

\section{INTRODUCTION}
\begin{figure}[t]
    \centering
    \includegraphics[width=1.0\linewidth]{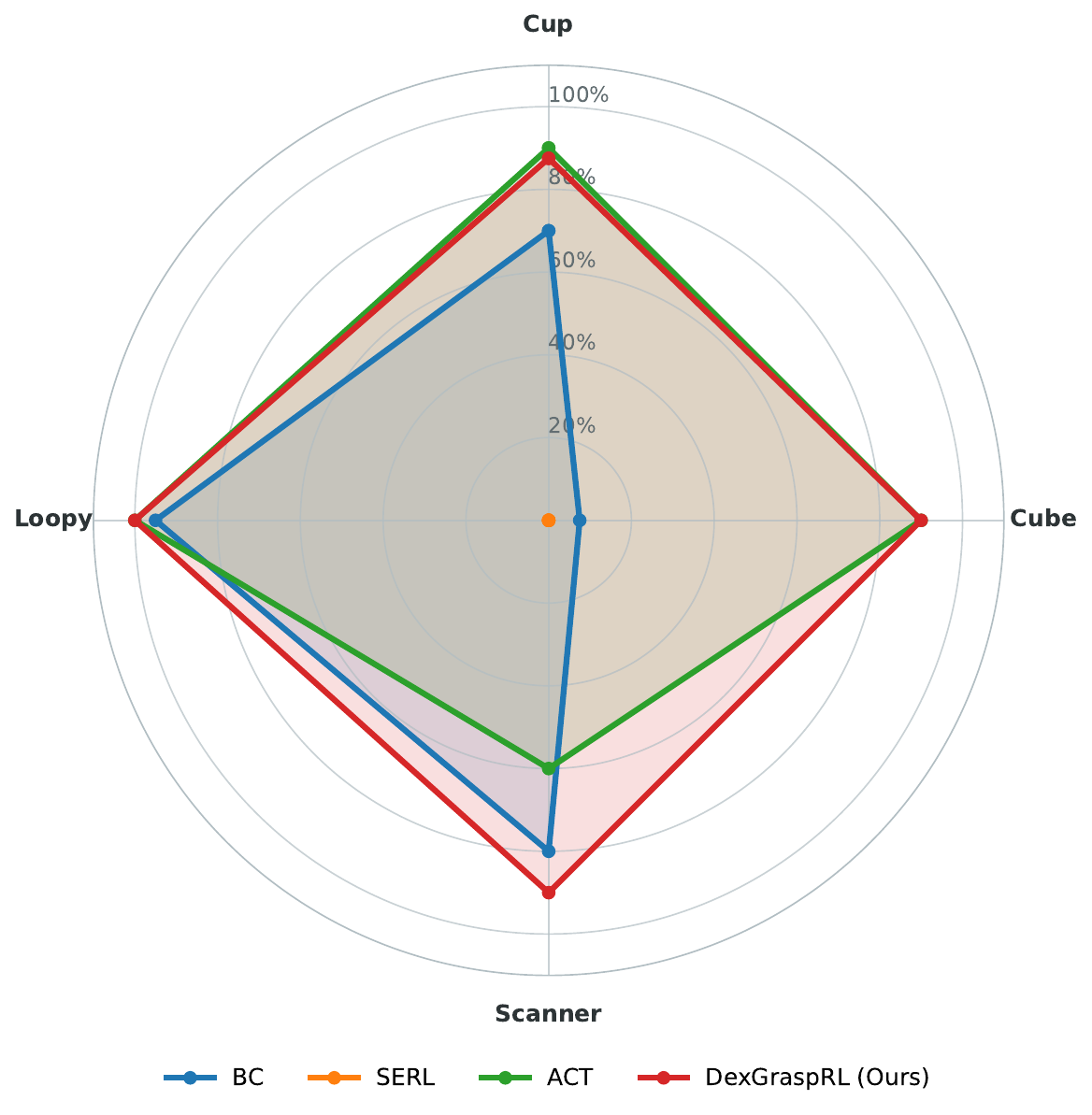}
    \caption{Performance comparision among BC, SERL, ACT, and our \textbf{DexGraspRL}, on real-world tasks.}
    \label{fig:radar}
\end{figure}
Dexterous grasping \cite{chen2022dextransferrealworldmultifingered, wan2023unidexgraspimprovingdexterousgrasping, wang2025unigrasptransformersimplifiedpolicydistillation, zhang2024dexgraspnet20learninggenerative, liu2024realdex}, which refers to the process of establishing a secure and stable coupling between the dexterous hand and the object, has been extensively investigated as a fundamental capability for dexterous manipulation.
Dexterous hands \cite{wang2024dexcap}, owing to their structural resemblance to human hands and their high degrees of freedom (DOFs), provide enhanced flexibility, precision, and adaptability in task execution. However, these advantages are accompanied by considerable challenges in developing effective grasping strategies and algorithms. 
In the realm of simpler end-effectors, such as grippers, analytical motion planning-based methods \cite{5652970} are widely employed for grasping and manipulation tasks in real-world scenarios. Nevertheless, these methods become impractical for dexterous hands due to the substantial search space resulting from their high DOFs \cite{overviewarticle}. Consequently, researchers are increasingly focusing on learning-based approaches \cite{chen2022dextransferrealworldmultifingered, wan2023unidexgraspimprovingdexterousgrasping, wang2025unigrasptransformersimplifiedpolicydistillation, zhang2024dexgraspnet20learninggenerative, liu2024realdex}, such as reinforcement learning (RL) \cite{mnih2013playingatarideepreinforcement} and imitation learning (IL) \cite{torabi2018behavioralcloningobservation}, demonstrating promising results in complex dexterous manipulation.

Research on learning-based methods for dexterous grasping primarily relies on two paradigms:
\textit{(i)} \textit{Cascade:} This paradigm \cite{wu2022learninggeneralizabledexterousmanipulation} first generates appropriate grasping poses for the target object and subsequently executes these poses through motion planning. These methods benefit from large-scale dataset pretraining, enabling them to effectively grasp a wide variety of objects across diverse scenarios.
\textit{(ii)} \textit{End-to-End:} This paradigm \cite{LI2024102792} engages directly in grasp planning through RL and IL without necessitating pre-grasp initialization. These methods utilize a streamlined architecture, mitigating issues related to miscalibration and inaccurate target grasp poses during the grasp generation phase, thus offering more robust and scalable solutions.
Building upon these paradigms, substantial algorithms have been proposed to achieve intuitive, adaptable, and human-like dexterous grasping.

Despite these advancements, dexterous grasping methods for real-world applications remain relatively scarce. Existing methods primarily utilize simulators for training; however, simulating the complexities of the real world presents significant challenges, resulting in an inevitable sim-to-real gap. Although IL learns directly from demonstrations, circumventing this gap, it often struggles to generalize beyond the training data. Furthermore, this sim-to-real gap is exacerbated by the impracticality of collecting expert trajectories and grasping poses across the diverse range of objects and environmental variations necessary for general grasping. 

\textit{"Can we learn dexterous grasping policies through performing reinforcement learning directly in the real world?"} Inspired by this natural thought, we explore the methods to bridge the sim-to-real gap for dexterous grasping tasks by employing real-world robotic reinforcement learning.
In contrast to simulations, RL in real-world scenarios must contend with challenges related to sample efficiency, autonomous and safe training, and reward specifications. To address these challenges, we propose \textbf{DexGraspRL}, a reinforcement learning framework specifically designed to directly acquire dexterous grasping skills in real-world environments.
\textbf{DexGraspRL} comprises two critical components that facilitate the stable and efficient learning of dexterous grasping policies:
\textit{(i)} a system that ensures automation and efficiency in real-world robotic reinforcement learning for dexterous grasping while maintaining compatibility across diverse tasks;
\textit{(ii)} a RL algorithm that guarantees sample-efficient learning from limited online real-world rollouts.
To specifically address the exploration challenges arising from the vast configuration space of dexterous hands and sparse rewards, we pretrain the policy with imitation learning prior to engaging in online RL. Furthermore, to mitigate the issue of catastrophic forgetting associated with an untrained critic, we have devised a regularization mechanism within this algorithm. This mechanism adaptively balances value maximization with the preservation of the pretrained policy.
As evidenced by our empirical results, meticulous coordination among these designs is essential for achieving optimal performance. As depicted in Fig.\ref{fig:radar}, our experiments in real-world scenarios demonstrate that \textbf{DexGraspRL} successfully accomplishes several tasks that traditional real-world robotic reinforcement learning methods struggle to complete. Furthermore, \textbf{DexGraspRL} outperforms IL baselines, attaining an average success rate of nearly 92\% and a reduced average cycle time.
In summary, the primary contributions of this paper are as follows:
\begin{itemize}
\item We have established a real-world reinforcement learning framework specifically designed for dexterous grasping, thereby paving the way for acquiring dexterous hand skills in the real world.
\item We propose an efficient real-world reinforcement learning algorithm that employs imitation learning to address the exploration and efficiency challenges posed by the vast configuration space of dexterous hands.
\item \textbf{DexGraspRL} successfully executes various real-world tasks characterized by diverse geometries and properties, revealing superior policies compared to the IL approach. This demonstrates the potential of RL beyond the capabilities of human experts.
\end{itemize}

\section{RELATED WORK}
\paragraph{Dexterous grasping} 
Current research addressing dexterous grasping tasks through learning-based methods \cite{chen2022dextransferrealworldmultifingered, wan2023unidexgraspimprovingdexterousgrasping, wang2025unigrasptransformersimplifiedpolicydistillation, zhang2024dexgraspnet20learninggenerative, liu2024realdex} can be categorized into two effective paradigms:
\textit{(i) Cascade:} Within this paradigm \cite{wu2022learninggeneralizabledexterousmanipulation}, dexterous grasping tasks are decoupled into two distinct subtasks: Grasp Generation (GG) \cite{wang2025unigrasptransformersimplifiedpolicydistillation} and Grasp Execution (GE). The GG phase generates appropriate, diverse, and affordance-aware grasp postures by training expressive generative models on large-scale datasets informed by various object properties and grasping intents. This approach facilitates adaptation to a broader range of grasping scenarios. Subsequently, the GE phase executes the targeted grasp poses through motion planning, which can be efficiently managed using RL algorithms. By leveraging pre-grasp initialization, the RL-based motion planning algorithms \cite{wang2025unigrasptransformersimplifiedpolicydistillation, zhang2024dexgraspnet20learninggenerative, liu2024realdex} can be simplified to Goal-Conditioned Reinforcement Learning (GCRL), resulting in reduced sample complexity. For the GG phase, extensive datasets comprising large-scale, high-quality, and diverse grasping poses for dexterous hands have been established. Building upon these datasets, substantial algorithms have been proposed to achieve intuitive, adaptable, and human-like dexterous grasping. Although promising diversity, the generated static grasp poses in the GG phase are often not validated in real-world scenarios, adversely affecting overall success.
\textit{(ii) End-to-End:} This paradigm \cite{LI2024102792} directly learns the entire grasping process through imitation learning and reinforcement learning. These methods offer more robust and scalable solutions by utilizing a streamlined architecture. While effective, these methods are either constrained by a limited number of expert demonstrations or suffer significant performance degradation when transferring the policy from simulation to reality.
To bridge the sim-to-real gap, we propose \textbf{DexGraspRL}, a real-world robotic reinforcement learning framework capable of end-to-end learning for dexterous grasping tasks in real-world environments.

\paragraph{Real-world robotic RL} 
Reinforcement learning in real-world environments presents significant challenges due to its demands for high sample efficiency, autonomous training, precise reward specifications, and safety considerations. Several methods have been proposed to address these issues, including off-policy methods \cite{luo2024serl, luo2024hilserl, smith2022walkparklearningwalk}, on-policy methods \cite{zhu2018dexterousmanipulationdeepreinforcement}, model-based methods \cite{wu2022daydreamerworldmodelsphysical, rafailov2020offlinereinforcementlearningimages, nagabandi2019deepdynamicsmodelslearning, 8594353}, and methods \cite{huang2025mentormixtureofexpertsnetworktaskoriented} that employ a mixture-of-experts structure combined with task-oriented perturbations to enhance the agent’s capability in managing complex tasks. However, these methods are primarily evaluated on gripper robots, failing to address complex end-effectors with intricate mechanical structures and high degrees of freedom, such as dexterous hands. Similar to our work, \cite{openai2019learningdexterousinhandmanipulation} succeeded in employing reinforcement learning to develop dexterous in-hand manipulation policies from visual inputs in real-world environments. Nevertheless, in-hand manipulation circumvents the challenges of motion planning, resulting in simplified settings. Consequently, this paper expands the boundaries of real-world robotic reinforcement learning to include dexterous grasping tasks.

% \paragraph{Bootstrapping RL with demonstrations} Leveraging demonstrations to extract useful skills and representations for RL has become widely adopted due to its ability to enhance sample efficiency \cite{huang2025dexteroushandmanipulationefficient}. Offline datasets can be classified into two categories: demonstrations and annotat1ed trajectories. Demonstrations, which consist of high-quality data devoid of associated rewards, are utilized to initialize the policy through imitation learning (IL). While annotated trajectories, which include reward annotations but may not be optimal, are employed to initialize both the policy and value function using offline RL. Compared to IL, offline RL presents a more advantageous option for initialization, as it pretrains all components necessary for online RL. Furthermore, in the presence of a reward signal, the policy pretrained with offline RL demonstrates superior performance due to the ability of Q-learning to prioritize critical decisions.

\begin{figure*}[!htb]
    \centering
    \includegraphics[width=1.0\linewidth]{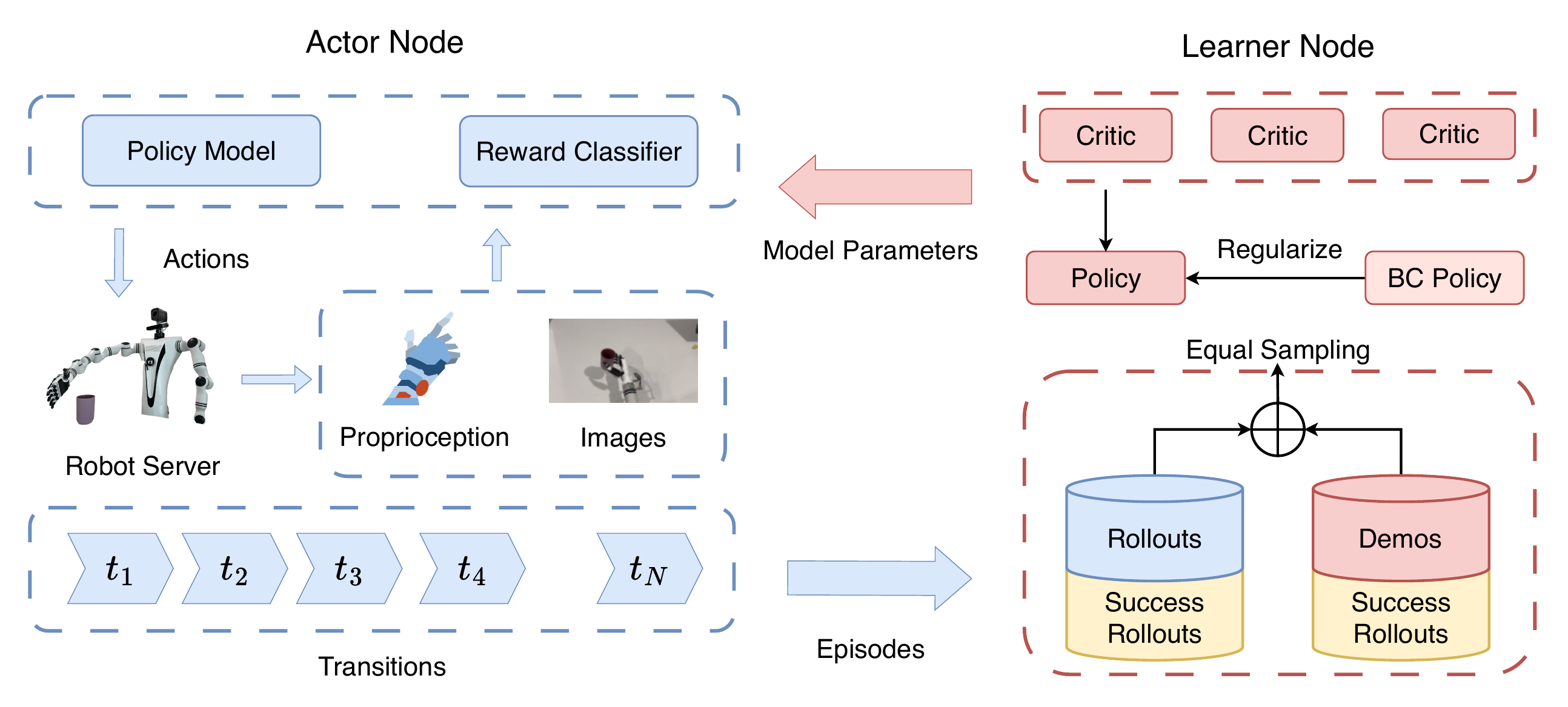}
    \caption{\textbf{Overview of DexGraspRL.} This figure illustrates the architecture of our system, which consists of two components: the actor and the learner. These components engage in asynchronous communication to enable efficient training.
    % The actor process receives updated policy parameters from the learner process, interacts with the environment, and transmits collected rollouts to the replay buffers. The learner process samples batches evenly from the two replay buffers and updates the policy. During the policy update, the ensemble of critics provides the reinforcement learning (RL) loss, while the pretrained policy regularizes the actor to prevent forgetting.
    }
    \label{fig:overview}
\end{figure*}

\section{PROBLEM FORMULATION}
In this work, we focus on developing a policy for dexterous grasping utilizing RGB observations and proprioceptive information within a tabletop setting. This problem can be formulated as a Markov Decision Process (MDP), wherein RL learns policies that maximize cumulative rewards in a specified environment. Formally, we consider an MDP defined by a tuple $(\mathcal{S}, \mathcal{A}, P, r, \gamma)$, where both state $\mathcal{S}$ and action spaces $\mathcal{A}$ are continuous. Here, $P(s'|s,a)$ denotes the state transition probability, $r(s,a)$ is the reward function, and $\gamma \in [0,1]$ is the discount factor. The goal of RL is to find a policy $\pi(a|s)$ that maximizes the expected \textit{return}, which is the sum of discounted future rewards as $G_t = \sum_{i=0}^{\infty} \gamma^i r_{t+i}.$

In contrast to simulators, where reward functions can be explicitly defined based on the underlying states, specifying reward functions presents a significant challenge for RL in real-world environments. Consequently, in our work, we only specified sparse rewards, assigning a reward of 1 when the dexterous grasping tasks are deemed successful, and a consistent reward of 0 otherwise.

\section{METHOD}
In this section, we first provide an overview of our \textbf{DexGraspRL} framework. We then present a comprehensive description of the online training system we developed, along with the RL algorithm we proposed. Finally, we include the pseudo-code for our method.

\subsection{Overview}
\textbf{System Design.} Our training system establishes a comprehensive pipeline encompassing expert demonstration collection, reward annotation, task status detection, robotic control, and an online training architecture. A failure in any component may lead to an underperforming policy. Specifically, we teleoperate the robotic hand to execute tasks successfully while recording each episode. Subsequently, we automatically replay these episodes and annotate the frames that indicate successful task completion. Utilizing the annotated rollouts, we train a binary classifier to determine whether a task has been performed successfully. This binary classifier functions as the reward detector during the reinforcement learning phase. Building upon prior efforts \cite{luo2024hilserl}, we implement our online training architecture. To facilitate efficient online reinforcement learning training, we separate the actor and learner into two distinct threads. The actor infers actions to interact with the robot controller and transmits the collected rollouts to the learner. Meanwhile, the learner updates the policy based on the provided rollouts and periodically refreshes the actor's policy. This distributed architecture not only decouples policy learning from environmental interactions but also maintains a consistent control frequency.

\textbf{Learning Algorithm.} To address the exploration and efficiency challenges inherent in RL for dexterous grasping, our algorithm comprises two stages: BC pretraining and Regularized RL finetuning. Initially, we pretrain the policy by replicating expert demonstrations. Subsequently, we finetune the pretrained policy using the modified RLPD algorithm \cite{ball2023efficientonlinereinforcementlearning}, which utilizes symmetric sampling between demonstrations and online rollouts. This approach has proven to be sample-efficient and reliable in real-world RL applications \cite{luo2024serl}. However, naive RL applied to a pretrained policy exhibits forgetting behavior, leading to performance degradation \cite{haldar2023watchmatchsuperchargingimitation}. To mitigate this challenge, we incorporate a regularization term into the policy objective to adaptively balance reinforcement learning with the pretrained policy.

\subsection{DexGraspRL System}

\textbf{Reward Specification with Binary Classifiers}
Reward specification through hand design is often infeasible, as certain forms of success status are challenging to define using explicit rules. This is particularly true for dexterous grasping tasks, where a successful grasp must be affordance-aware, task-specific, and occasionally contingent upon human preference. Formulating an explicit rule to ascertain a successful grasp based on image observations is clearly intractable.
However, the features that determine a successful grasp can be effectively captured by neural networks, provided that sufficient training data are available. 
Consequently, we train binary classifiers using the annotated rollouts to predict rewards based on image observations. Although binary rewards may face challenges related to sparsity, they can be easily adapted to various tasks, thereby circumventing the substantial effort required to design reward functions.

\textbf{High-performance Robot Controller}
We specifically implement a high-performance robot controller that utilizes gRPC rather than HTTP, distinguishing it from conventional practices. In comparison to controllers built on HTTP, our controller demonstrates reduced latency and improved stability in receiving actions and returning observations.

\textbf{Distributed Architecture}
We segregate the actor and learner into two distinct threads. The learner manages both the online replay buffer and the demo replay buffer, sampling batches from these buffers to update the policy. The actor supervises the robot controller, inferring actions from the current observations and executing them accordingly. At each step, the current rollout is sent to the online replay buffer. Importantly, when an episode is classified as successful, it is regarded as a high-quality episode and subsequently transferred to the demo replay buffer. Throughout this process, the learner updates the policy based on the available data and periodically refreshes the actor's policy.

\subsection{DexGraspRL Algorithm}
\textbf{BC Pretraining}
RL from scratch in real-world environments faces challenges related to exploration and sample efficiency \cite{johannink2018residualreinforcementlearningrobot}. To address these issues, we pretrained the policy through replicating expert demonstrations \cite{torabi2018behavioralcloningobservation}. Specifically, we collected 30 demonstrations via teleoperation and employed behavior cloning with the following objective:
\begin{equation}
    \mathcal{L}^{BC}=\mathbb{E}_{(s^e,a^e)\thicksim\mathcal{\tau}^e}\|a^e-\pi^{BC}(s^e)\|^2
\end{equation}
where, $\mathcal{\tau}^e$ represents expert demonstrations, $s^e$ and $a^e$ denote the corresponding state and action in expert demonstrations, and $\pi^{BC}$ signifies the learned behavior cloning policy. 

\begin{algorithm}[!t]
  \caption{DexGraspRL}
  \begin{algorithmic}[1]
    \STATE \textbf{Initialization:} Randomly initialize Critic parameters $\theta_i$ (set target parameters $\theta_i' = \theta_i$) for $i = 1, 2, \ldots, E$.
    \STATE  \textcolor{red}{Load pretrained Actor parameters $\phi'$, and initialize Actor parameters $\phi = \phi'$.}
    \STATE Set the discount factor $\gamma$, temperature $\alpha$, ensemble size $E$, UTD ratio $G$, and critic exponential moving average (EMA) weight $\rho$.
    \STATE Determine the number of Critic targets to subset, $Z \in \{1, 2\}$.
    \STATE Initialize an empty replay buffer $\mathcal{R}$.
    \STATE Initialize the buffer $\mathcal{D}$ with offline data.
    
    \WHILE{True}
        \FOR{$g = 1, \ldots, G$}
            \STATE Sample a minibatch $b_R$ of size $\frac{N}{2}$ from $\mathcal{R}$.
            \STATE Sample a minibatch $b_D$ of size $\frac{N}{2}$ from $\mathcal{D}$.
            \STATE Combine $b_R$ and $b_D$ to form a batch $b$ of size $N$.
            \STATE Sample a set $\mathcal{Z}$ of $Z$ indices from $\{1, 2, \ldots, E\}$.
            
            \STATE \textbf{Compute target:}
            \vspace{-2mm}
            \[
            y = r + \gamma \left( \min_{i \in \mathcal{Z}} Q_{\theta_i'}(s', \tilde{a}') \right), \quad \tilde{a}' \sim \pi_\phi(\cdot | s').
            \]
            \vspace{-2mm}
            
            \FOR{$i = 1, \ldots, E$}
                \STATE Update $\theta_i$ by minimizing the loss:
                \vspace{-2mm}
                \[
                L = \frac{1}{N} \sum_{j=1}^{N} (y - Q_{\theta_i}(s, a))^2.
                \]
                \vspace{-4mm}
            \ENDFOR
            
            \STATE Update target networks: $\theta_i' \leftarrow \rho \theta_i' + (1 - \rho) \theta_i$.
        \ENDFOR
        
        \STATE Sample a minibatch $b_R$ of size $\frac{N}{2}$ from $\mathcal{R}$.
        \STATE Sample a minibatch $b_D$ of size $\frac{N}{2}$ from $\mathcal{D}$.
        \STATE Combine $b_R$ and $b_D$ to form a batch $b$ of size $N$.
        
        \STATE \textcolor{blue}{\textbf{For batch $b_R$, compute:}
        \[
        \lambda(\phi) = \frac{1}{N} \sum_{j=1}^{N} \frac{1}{E} \sum_{i=1}^{E} \mathbb{I}_{Q_{\theta_{i}}(s, \pi_{\phi'}(s)) > Q_{\theta_{i}}(s, \pi_{\phi}(s))}.
        \]
        }
        
        \STATE \textcolor{blue}{\textbf{For batch $b_D$, compute:}
        \[
        \mathcal{L}^{BC} = \frac{2}{N} \sum_{i = 1}^{\frac{N}{2}} \beta \lambda(\phi) \|a^e - \pi(s^e)\|^2.
        \]
        }
        \STATE \textbf{Update $\phi$ by maximizing the objective:}
        \vspace{-2mm}
        \[
        \frac{1}{E} \sum_{i = 1}^{E} Q_{\theta_{i}}(s, \tilde{a}) - \alpha \log \pi_{\phi}(\tilde{a} | s) \textcolor{blue}{+ \mathcal{L}^{BC}}, \quad \tilde{a} \sim \pi_{\phi}(\cdot | s).
        \]
        \vspace{-3mm}
    \ENDWHILE
  \end{algorithmic}
  \label{alg:DexGraspRL}
\end{algorithm}

\textbf{Regularized RL Finetuning}
Although RL from a pretrained policy can significantly alleviate exploration issues, utilizing reinforcement learning with an untrained critic may lead to catastrophic forgetting\cite{nair2021awacacceleratingonlinereinforcement, uchendu2023jumpstartreinforcementlearning}. As seen in Eq.\eqref{eq:RL_loss}, we address this issue by incorporating a regularization term to balance the exploration of novel behaviors and the exploitation of expert behaviors. 
\begin{equation}
\label{eq:RL_loss}
    \mathcal{L}^{RL}=(1-\lambda(\pi))\mathbb{E}_{(s,a)\sim\mathcal{D}_{\beta}}[Q(s,a)] \\
    -\alpha\lambda(\pi)\mathbb{E}_{(s^e,a^e)\sim\mathcal{T}^e}\|a^e-\pi(s^e)\|^2.
\end{equation}
Here, \( Q(s, a) \) denotes the Q-value derived from the critic. The parameter \( \alpha \) represents a fixed weight, while \( \lambda(\pi) \) is a policy-dependent adaptive weight that regulates the contributions of the two loss terms. \( \mathcal{D}_{\beta} \) refers to the online replay buffer.

Since the balancing coefficient \( \lambda(\pi) \) between value maximization and regularization significantly influences the algorithm's performance, it is essential to develop a principled method for its determination. Built upon previous studies \cite{rajeswaran2018learningcomplexdexterousmanipulation, jena2020augmentinggailbcsample, nguyen2022leveragingfullyobservablepolicies}, we employ an adaptive scheme to ascertain \( \lambda(\pi) \):
\begin{equation}
\lambda(\pi)=\mathbb{E}_{(s,\cdot)\sim\mathcal{D}_\beta}\left[\mathbb{I}_{Q(s,\pi^{BC}(s))>Q(s,\pi(s))}\right].
\end{equation}
Here, $\lambda(\pi)$ denotes the degree to which the pretrained policy surpasses the current policy. A higher value of $\lambda(\pi)$ indicates that greater effort should be devoted to regularization.

\subsection{Training Pipeline}
In Algorithm \ref{alg:DexGraspRL}, we present the pseudo-code for our method. The elements highlighted in \textcolor{red}{red} indicate the loading of the pretrained actor parameters at the outset of the reinforcement learning phase, while the lines in \textcolor{blue}{blue} illustrate the regularization mechanism.
The key factors of our algorithm can be summarized as follows:
\begin{itemize}
\item {\em Line 2:} Pretrain the policy by replicating expert demonstrations.
\item {\em Lines 9-10 and 19-21:} Employ a symmetric sampling approach to integrate online and demonstration rollouts.
\item {\em Lines 8-18:} Implement high UTD updates to ensure optimal efficiency.
\item {\em Lines 22-24:} Apply regularization using the pretrained policy.
\end{itemize}

\begin{table*}[bp]
\centering
\small  % Reduce font size for double-column format
\scalebox{1.2}{
\begin{tabular}{@{}c c c c c c@{}}
\toprule
\textbf{Tasks} & \textbf{Demos} & \textbf{Image Inputs} & \textbf{Reset} & \textbf{Training Time} & \textbf{Classifier} \\
\midrule
Cup     & 40 & Chest Camera, Head Camera & Fixed & 59 mins & 98\% \\
Cube     & 36 & Chest Camera, Head Camera & Fixed & 72 mins & 96\% \\
Scanner  & 31 & Chest Camera, Head Camera & Random & 41 mins & 95\% \\
Loopy   & 36 & Chest Camera, Head Camera & Random & 42 mins & 99\% \\
\bottomrule
\end{tabular}
}
\vspace{0.5em}
\captionsetup{labelfont=bf, skip=6pt}
\caption{Task Parameters: The initial position of the target object is randomly reset within a specified region.}
\label{tb:task_parameters}
\end{table*}

\section{EXPERIMENTS}
In this section, we discuss our experiments aimed at evaluating the \textbf{DexGraspRL} framework. We first outline our experimental settings. We then present our experimental results and conduct an ablation analysis to investigate the effects of various components.

\subsection{Experimental Settings}
\textbf{Hardware}
In our experiments, we deploy our policies on the Inspire robot, which includes a dexterous hand with 6 degrees of freedom, to perform various dexterous grasping tasks using its right arm and hand. For all tasks, the observation space consists of camera images captured from the robot's chest and head, along with proprioceptive information, such as the end-effector pose and the angles of the robot's fingers. The action space corresponds to the proprioceptive information found in the observation space. Specifically, the end-effector pose is represented as the relative position with respect to its initial position at the beginning of the task, comprising 3D Cartesian displacement and 3D rotational deviation using axis-angle representation.

\begin{figure}[!htb]
    \centering
    \includegraphics[width=1.0\linewidth]{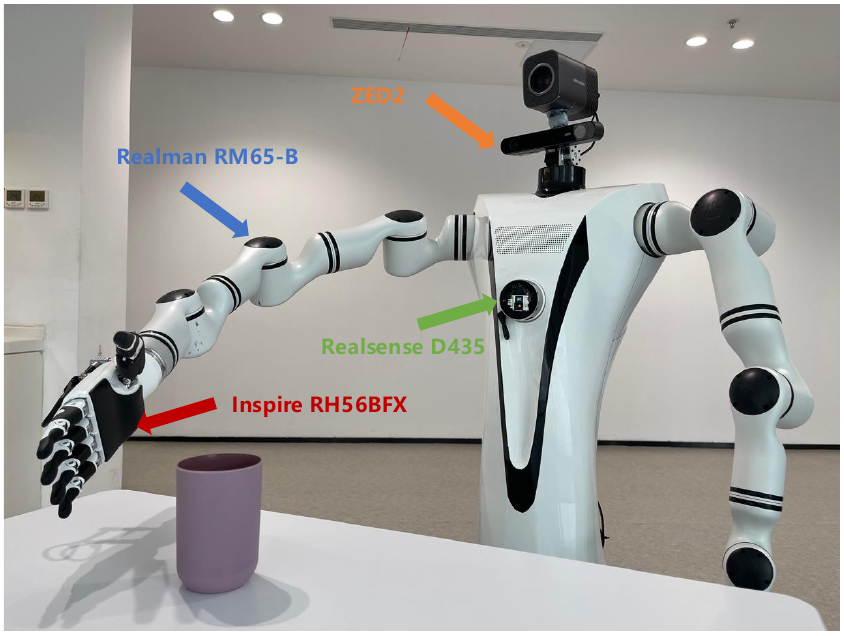}
    \caption{Illustration of the robot used in our experiments. The robot is tasked with grasping a cup, utilizing image observations from the ZED2 and Realsense D435 cameras as inputs to control its arm and hand for task completion.}
    \label{fig:formulation}
\end{figure}

\textbf{Data Collection}
Unlike other robotic systems, teleoperation via keyboard is impractical. We control the robotic hand using electromagnetic field (EMF) gloves \cite{wang2024dexcapscalableportablemocap} that track finger movements and monitor the six degrees of freedom (6-DoF) of the wrist pose based on the SLAM algorithm. This teleoperation device enables us to manipulate the robotic hand as effortlessly as we do our own, thereby facilitating task execution. During task execution, we record camera images captured from the robot's chest and head, along with the end-effector pose and the angles of the robot's fingers. These data serve as expert demonstrations and constitute the training data for the reward classifier after being annotated with our automatic scripts.

% During the policy deployment phase, we trained a binary classifier to serve as a reward detector. This classifier analyzes images to predict whether the current state successfully accomplishes the task. When the reward detector identifies the current state as successful, the episode concludes, and we manually reset the environment.

\begin{figure*}[htb]
    \centering
    \includegraphics[width=1.0\linewidth]{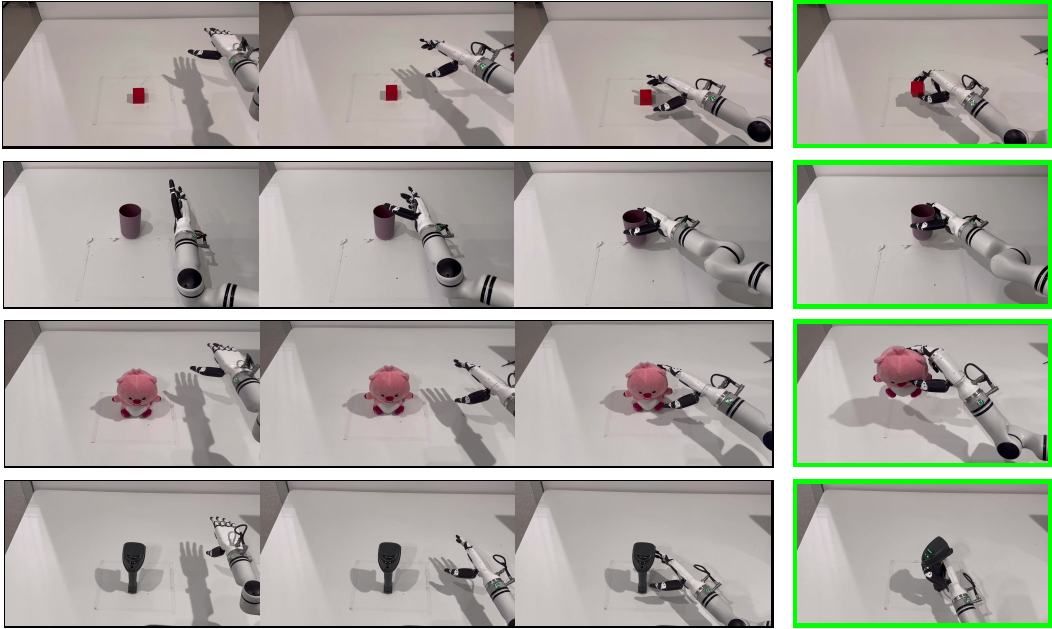}
    \caption{Illustration of the robot performing each task with our method. For each task, the robot starts from a predefined initial pose and subsequently performs the task by utilizing the proprioceptive state and image observations. The process continues until the robot receives a reward for successfully completing the task, as indicated within the green box.}
    \label{fig:tasks}
\end{figure*}

\textbf{Task Settings}
We present the visualizations of the tasks in Figure \ref{fig:tasks}, along with detailed descriptions provided below:

\begin{itemize}
\item \textbf{Cup:} In this task, the robot is required to grasp and lift a cup from a designated position on a table. This task assesses the robot's precision in handling lightweight objects and its ability to adjust grasp strength.
\item \textbf{Scanner:} In this task, the robot must pick up a scanner using an affordance-aware pose. This task evaluates the robot's capability to grasp objects with complex geometries and its adaptability in mastering the affordance-aware grasping technique.
\item \textbf{Cube:} In this task, the robot is tasked with pinching and lifting a cube with its fingers. This task assesses the robot's dexterity in finger control.
\item \textbf{Loopy:} The robot is tasked with grasping and lifting a loopy doll. This endeavor presents unique challenges due to the doll's unconventional shape and the characteristics of the soft materials used.
\end{itemize}

\textbf{Baselines}
To illustrate the superiority of \textbf{DexGraspRL}, we compare it with the following baselines:
\begin{itemize}
    \item \textbf{BC:} Behavior Cloning (BC) is a common IL method in which an agent learns to replicate an expert's behavior by training on expert demonstrations.
    \item \textbf{SERL:} Sample-Efficient Robotic Reinforcement Learning (SERL) is an open-source software framework that provides a high-quality RL implementation specifically designed for real-world robotic learning.
    \item \textbf{DexGraspRL$^\dagger$:} DexGraspRL$^\dagger$ is a ablated version of DexGraspRL, where DexGraspRL$^\dagger$ directly applies RL to the pretrained policy without any regularization.
    \item \textbf{ACT:} Action Chunking with Transformers (ACT) is the most advanced IL method, which learns a generative model over action sequences.
\end{itemize}

\begin{table*}[htb]
\centering
\small  % 缩小字号以适应双栏
\scalebox{1.2}{
\begin{tabular}{@{}l *{5}{S[table-format=2.1]@{\hspace{1em}}S[table-format=-2.2]@{\hspace{1em}}}@{}}
\toprule
\multirow{2}{*}{Method} & 
\multicolumn{2}{c}{\textbf{Cup}} & 
\multicolumn{2}{c}{\textbf{Cube}} & 
\multicolumn{2}{c}{\textbf{Scanner}} & 
\multicolumn{2}{c}{\textbf{Loopy}} &
\multicolumn{2}{c}{\textbf{Average}} \\
\cmidrule(lr){2-3} \cmidrule(lr){4-5} \cmidrule(lr){6-7} \cmidrule(lr){8-9} \cmidrule(lr){10-11}
& {SR} & {CT} & {SR } & {CT} & {SR } & {CT} & {SR } & {CT} & {SR } & {CT}  \\
\midrule
BC        & 70.0 & 122.21 & 7.5 & 129.67 & 80.0 & 204.25 & 95.0 & 116.11 & 61.4 & 143.1 \\
SERL      & 0 & N/A & 0 & N/A & 0 & N/A & 0  & N/A & 0  & N/A \\
DexGraspRL$^\dagger$ & 0 & N/A & 0 & N/A & 0 & N/A & \textbf{100.0} & \textbf{101.8} & 25.0 & N/A  \\
\textbf{DexGraspRL (Ours)}  & \textbf{87.5} & \textbf{68.51} & \textbf{90.0} & \textbf{110.55} & \textbf{90.0} & \textbf{147.19} & \textbf{100.0} & 115.55 & \textbf{91.9} & \textbf{110.45}\\
\bottomrule
\end{tabular}
}
\captionsetup{labelfont=bf, skip=6pt}
\caption{Experimental results comparing our method with baselines. {\scriptsize$\dagger$} Ablated version without regularization terms. SR: Success Rate, CT: Total steps for complete the task.}
\label{tb:results}
\end{table*}

\subsection{Experimental Results}
Now, we discuss our empirical results for the real-world evaluation. For each task, we report success rate, CT and detailed task parameters. The training time includes all scripted motion, policy rollouts, intended stops, as well as onboard computation which is carried on a single NVIDIA RTX 4090 GPU. Unless otherwise noted, all results are based on 40 evaluation trials. 
% \begin{table*}[htb]
% \centering
% \small  % 缩小字号以适应双栏
% \scalebox{1.2}{
% \begin{tabular}{@{}l *{5}{S[table-format=2.1]@{\hspace{1em}}S[table-format=-2.2]@{\hspace{1em}}}@{}}
% \toprule
% \multirow{2}{*}{Method} & 
% \multicolumn{2}{c}{\textbf{Cup}} & 
% \multicolumn{2}{c}{\textbf{Cube}} & 
% \multicolumn{2}{c}{\textbf{Scanner}} & 
% \multicolumn{2}{c}{\textbf{Loopy}} \\
% \cmidrule(lr){2-3} \cmidrule(lr){4-5} \cmidrule(lr){6-7} \cmidrule(lr){8-9}
% & {SR} & {CT} & {SR } & {CT} & {SR } & {CT} & {SR } & {CT} \\
% \midrule
% ACT       &  \textbf{90.0} & 102.33 & \textbf{90.0} & \textbf{91.11} & \textbf{60.0} & \textbf{86.17} & \textbf{100.0} & \textbf{85.4} \\
% BC        & 70.0 & 122.21 & 7.5 & 129.67 & 80.0 & 204.25 & 95.0 & 116.11 \\
% SERL      & 0 & N/A & 0 & N/A & 0 & N/A & 0  & N/A \\
% DexGraspRL$^\dagger$ & 0 & N/A & 0 & N/A & 0 & N/A & 100.0 & \textbf{101.8} \\
% \textbf{DexGraspRL (Ours)}  & \textbf{87.5} & \textbf{68.51} & \textbf{90.0} & \textbf{110.55} & \textbf{90.0} & \textbf{147.19} & \textbf{100.0} & 115.55  \\
% \bottomrule
% \end{tabular}
% }
% \captionsetup{labelfont=bf, skip=6pt}
% \caption{Experimental results comparing our method with baselines. {\scriptsize$\dagger$} Ablated version without regularization terms. SR: Success Rate, CT: Total steps for complete the task.}
% \label{tb:results}
% \end{table*}

\textbf{Results.} As illustrated in Table.\ref{tb:results}, Our approach achieve competitive performance across all tasks mentioned above. Compared to BC, which serves as the initial policy for \textbf{DexGraspRL}, our approach outperforms it by a significant margin, particularly in the \textit{Cup} and \textit{Cube} task. In the \textit{Cup} task, \textbf{DexGraspRL} not only increases the success rate from 70\% to 87.5\% but also reduces the CT required to complete the task by half. This indicates that the policy can be further optimized beyond expert performance through efficient exploration in reinforcement learning. Meanwhile, in the \textit{Cube} task, which requires precise finger control, BC achieves only a 7.5\% success rate and an average CT of 129.67. This highlights that naive imitation learning struggles to master precise manipulation tasks due to compounding error issues. In contrast, \textbf{DexGraspRL} effectively masters the \textit{Cube} task, achieving a 90\% success rate and an average CT of 110.55 through extensive interactions with the real-world environment. This outcome demonstrates the superiority of real-world robotic reinforcement learning, as it can progressively refine its policy based on environmental feedback. Conversely, the SERL method, which focuses on real-world robotic reinforcement learning, struggles to achieve task completion due to the vast exploration space resulting from the large action space of dexterous hands and the extensive observation space derived from high-dimensional images. While \textbf{DexGraspRL} benefits from BC pretraining, successfully mastering dexterous hand grasping skills, SERL faces significant challenges in this regard.

\textbf{Ablation Analysis.} As illustrated in Table \ref{tb:results}, without regularization, DexGraspRL$^\dagger$ fails to complete three tasks during real-world training. We observe that RL encounters challenges in mastering precise and dexterous hand movements, such as grasping the cup, pinching the cube, and grabbing the scanner. Specifically, during the real-world RL training, the robotic hand tends to exhibit instability, causing the cup to be displaced and the scanner to be toppled. Consequently, the robot is unable to complete the tasks, resulting in a lack of reward signals and leading to persistent underperformance. In contrast, the \textit{Loopy} task, which does not require precise and intricate hand movements, is successfully accomplished by DexGraspRL$^\dagger$.
Compare DexGraspRL$^\dagger$ to BC, we observe a rapid decline in performance following RL fine-tuning without regularization, resulting in an average degradation of 36.4\% in the success rate. This highlights the importance of a regularization mechanism. Compare SERL to DexGraspRL$^\dagger$ and \textbf{DexGraspRL}, we observe that while SERL demonstrates impressive performance in gripper robots, it is ineffective in learning tasks without prior pretraining through IL.

\subsection{Discussions}
\textbf{Comparison with ACT.} We compare our method with Action Chunking with Transformers (ACT) to further demonstrate its superiority. ACT is a notable IL algorithm that addresses compounding error issues in IL by employing the transformer architecture, action chunking, and temporal ensemble techniques, resulting in significant performance improvements.
% \begin{figure}[!htb]
%     \centering
%     \includegraphics[width=1.0\linewidth]{assets/act.png}
%     \caption{Comparison with ACT. In our evaluation, we conducted 40 trials per task, recording both the success rate and the average number of steps taken by the robot to achieve success in each task.}
%     \label{fig:act}
% \end{figure}
\begin{table}[htb]
\centering
\small
\scalebox{1}{
\begin{tabular}{@{}l *{3}{S[table-format=2.1]@{\hspace{1em}}S[table-format=-2.2]@{\hspace{1em}}}@{}}
\toprule
\multirow{2}{*}{Tasks} &
\multicolumn{2}{c}{BC} &
\multicolumn{2}{c}{ACT} &
\multicolumn{2}{c}{\textbf{DexGraspRL}} \\
\cmidrule(lr){2-3} \cmidrule(lr){4-5} \cmidrule(lr){6-7}
& {SR} & {CT} & {SR} & {CT} & {SR} & {CT} \\
\midrule
Cup & 70.0 & 122.21 & \textbf{90.0} & 102.33 & 87.5 & \textbf{68.51} \\
Cube & 7.5 & 129.67 & \textbf{90.0} & \textbf{91.11} & \textbf{90.0} & 110.55 \\
Scanner & 80.0 & 204.25 & 60.0 & \textbf{86.17} & \textbf{90.0} & 147.19 \\
Loopy & 95.0 & 116.11 & \textbf{100.0} & \textbf{85.4} & \textbf{100.0} & 115.55 \\
Average & 61.4 & 143.1 & 85.0 & \textbf{91.25} & \textbf{91.9} & 110.45 \\
\bottomrule
\end{tabular}
}
\captionsetup{labelfont=bf, skip=6pt}
\caption{Comparison results with ACT.}
\label{tb:act}
\end{table}

As depicted in Table.\ref{tb:act}, \textbf{DexGraspRL} demonstrates competitive performance compared to the state-of-the-art IL algorithm, ACT. Notably, in the \textit{Cup} task, \textbf{DexGraspRL} accomplishes half the total steps of ACT while maintaining a comparable success rate. This indicates that RL can surpass the limitations of expert demonstrations to derive a more optimal policy through adequate trial and error. Conversely, in the \textit{Cube} task, although \textbf{DexGraspRL} achieves the same success rate, it is slower than ACT by an average of 20 steps. We attribute this issue to the underperforming backbone utilized in our method. In both \textbf{DexGraspRL} and BC, we employ an ImageNet pre-trained ResNet-10 as the vision backbone for the policy network. In contrast, ACT utilizes the most advanced transformer architecture. When comparing BC to ACT, it is evident that although both ACT and BC share the same objective, they yield completely different performance outcomes due to the differing backbones.

\begin{figure}[!htb]
    \centering
    \includegraphics[width=1.0\linewidth]{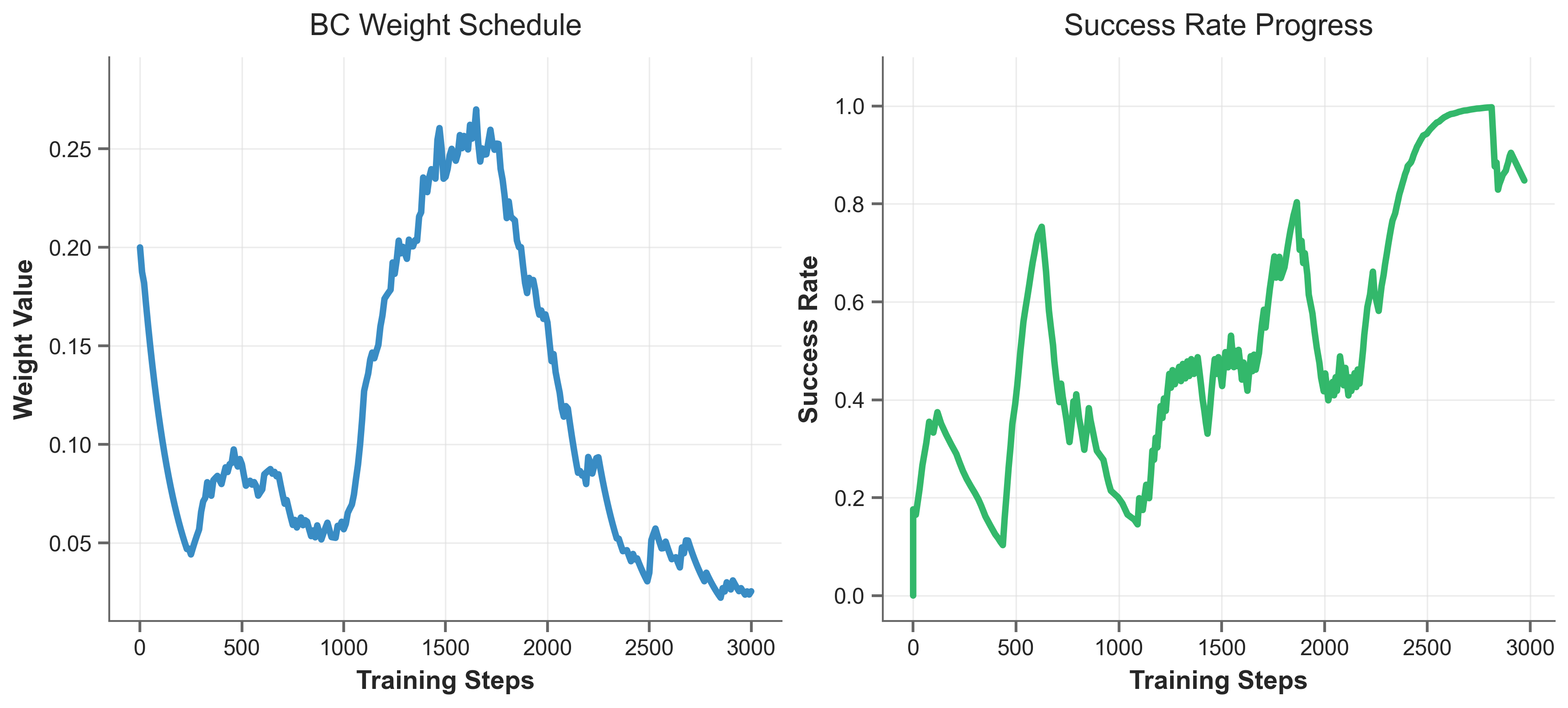}
    \caption{Regularization Analysis. This figure illustrates the changing trend of the balancing coefficient \( \lambda(\pi) \) and the success rate throughout the training process. A higher value of \textit{BC weight} indicates a greater extent to which the pretrained policy outperforms the current policy.}
    \label{fig:regularization}
\end{figure}

\textbf{Regularization Analysis.} To demonstrate how the regularization mechanism functions within the RL finetuning, we present Fig. \ref{fig:regularization}, illustrating the changing trends of the balancing coefficient and the success rate. Based on this figure, we analyze the relationship between these two metrics. 

As illustrated in Fig. \ref{fig:regularization}, the \textit{BC weight} shows no significant changes during the first 1000 steps, while the success rate remains consistently low. This is because the critic estimates the current policy based on online data, which must be sufficiently collected to fully demonstrate the performance of the current policy. After the first 1000 steps, the collected online data becomes sufficient to reflect the performance of the current policy. Consequently, the \textit{BC weight} exhibits a rapid increase, and the regularization begins to take effect. As the regularization progresses, the success rate increases, resulting in a decrease of the \textit{BC weight}. Finally, the success rate converges to nearly 100\%, while the \( \text{BC weight} \) declines to approximately zero.

\section{CONCLUSIONS}
This paper introduces \textbf{DexGraspRL}, a reinforcement learning framework that directly acquires dexterous grasping skills in real-world environments. In terms of system design, \textbf{DexGraspRL} establishes a framework that ensures the automation, efficiency, and compatibility of online RL training. Regarding the algorithm, \textbf{DexGraspRL} utilizes BC  to pretrain the policy, thereby mitigating the exploration challenges arising from the vast configuration space of dexterous hands. Furthermore, to address the challenges associated with catastrophic forgetting in RL fine-tuning, we propose an adaptive regularization mechanism. Our empirical results across several real-world dexterous grasping tasks demonstrate that \textbf{DexGraspRL} surpasses existing methods, achieving a 1.6-fold improvement in success rate and enhanced cycle time efficiency. Additionally, \textbf{DexGraspRL} successfully masters several tasks that current real-world robotic reinforcement learning methods struggle to accomplish.

Despite \textbf{DexGraspRL}'s strong performance and efficiency, we recognize that its effectiveness is significantly contingent upon the quality of the pretrained policy. In scenarios where the pretrained policy is inadequately trained, the exploration space becomes excessively vast, complicating the effective identification of sparse reward signals for reinforcement learning and thereby hindering the development of a practical policy. Therefore, our future research will concentrate on employing additional priors to enhance the sample efficiency of real-world RL training.

\addtolength{\textheight}{-12cm}  

%%%%%%%%%%%%%%%%%%%%%%%%%%%%%%%%%%%%%%%%%%%%%%%%%%%%%%%%%%%%%%%%%%%%%%%%%%%%%%%%

%%%%%%%%%%%%%%%%%%%%%%%%%%%%%%%%%%%%%%%%%%%%%%%%%%%%%%%%%%%%%%%%%%%%%%%%%%%%%%%%

%%%%%%%%%%%%%%%%%%%%%%%%%%%%%%%%%%%%%%%%%%%%%%%%%%%%%%%%%%%%%%%%%%%%%%%%%%%%%%%%
% \section*{APPENDIX}

% \section*{ACKNOWLEDGMENT}

%%%%%%%%%%%%%%%%%%%%%%%%%%%%%%%%%%%%%%%%%%%%%%%%%%%%%%%%%%%%%%%%%%%%%%%%%%%%%%%%

% \begin{thebibliography}{99}
% \end{thebibliography}

\bibliographystyle{IEEEtran}

\bibliography{root}{}

\end{document}